\documentclass{article}

\usepackage{microtype}
\usepackage{graphicx}
\usepackage{subfigure}
\usepackage{booktabs} 
\usepackage{amsmath}
\usepackage{amssymb}
\usepackage[algo2e]{algorithm2e}

\usepackage{hyperref}

\usepackage[accepted]{icml2019}

\icmltitlerunning{PDP: A General Neural Framework for Learning Constraint Satisfaction Solvers}

\begin{document}

\twocolumn[
\icmltitle{PDP: A General Neural Framework\\for Learning Constraint Satisfaction Solvers}



\icmlsetsymbol{equal}{*}

\begin{icmlauthorlist}
\icmlauthor{Saeed Amizadeh}{ms}
\icmlauthor{Sergiy Matusevych}{ms}
\icmlauthor{Markus Weimer}{ms}
\end{icmlauthorlist}

\icmlaffiliation{ms}{Microsoft, Redmond, WA, USA}

\icmlcorrespondingauthor{Saeed Amizadeh}{saamizad@microsoft.com}
\icmlcorrespondingauthor{Sergiy Matusevych}{sergiym@microsoft.com}

\icmlkeywords{Neuro-Symbolic Methods, Neural Combinatorial Optimization, Boolean Satisfiability}

\vskip 0.3in
]



\printAffiliationsAndNotice 

\begin{abstract}
There have been recent efforts for incorporating Graph Neural Network models for learning full-stack solvers for constraint satisfaction problems (CSP) and particularly Boolean satisfiability (SAT). Despite the unique representational power of these neural embedding models, it is not clear how the search strategy in the learned models actually works. On the other hand, by fixing the search strategy (e.g. greedy search), we would effectively deprive the neural models of learning better strategies than those given. In this paper, we propose a generic neural framework for learning CSP solvers that can be described in terms of probabilistic inference and yet learn search strategies beyond greedy search. Our framework is based on the idea of propagation, decimation and prediction (and hence the name PDP) in graphical models, and can be trained directly toward solving CSP in a fully unsupervised manner via energy minimization, as shown in the paper. Our experimental results demonstrate the effectiveness of our framework for SAT solving compared to both neural and the state-of-the-art baselines.           
\end{abstract}

\section{Introduction} \label{sec:intro}

Constraint satisfaction problems (CSP) \cite{kumar1992algorithms} constitute the cornerstone of combinatorial optimization in Computer Science. Boolean Satisfiability (SAT), in particular, is the most fundamental NP-complete problem in Computer Science with a wide range of applications from verification to planning and scheduling. There have been huge efforts in Computer Science \cite{biere2009handbook, knuth1997art} as well as Physics and Information Theory \cite{mezard2009information} to both understand the theoretical aspects of SAT and develop efficient search algorithms to solve it. 

While researchers have deeply studied different classes of the SAT problems and their properties -- e.g. \cite{nudelman2004understanding, krzakala2007gibbs, ansotegui2008measuring, ansotegui2012community}, most classical search algorithms and heuristics either are completely oblivious to the distribution of the input SAT problems or address rather a broad category of SAT problems -- e.g. \cite{jordi2015classification, newsham2014impact}. In many real scenarios, however, the input problem instances come from a quite narrow, domain-specific distribution. The common wisdom tells us in such cases, a domain-specific algorithm should do better than a general-purpose method. But even if we could develop such domain-specific algorithms for every single domain, the number of algorithms to select from for new instances would explode! In other words, having a \textit{generic} methodology seems to be mutually exclusive with being efficiently customized for every specific domain. To resolve this dilemma, Machine Learning comes into the equation. In particular, the main motivation is that by incorporating Machine Learning, we will have \textit{one} generic solver that can be specialized for specific domains based on data. This is a very attractive idea because we will have better solvers for specific problem domains and yet we will not have to deal with selecting between a myriad of solvers (which is a problem on its own -- e.g. \cite{xu2008satzilla}) for every new domain/instance. Obviously, we are not there yet, but the hope is by pursuing this direction, we can eventually come at such data-driven yet generic solvers.

In that vein, Machine Learning has been used for different aspects of CSP and SAT solving, from branch prediction (e.g. \cite{liang2016learning}) to algorithm and hyper-parameter selection -- e.g. \cite{xu2008satzilla, hutter2011sequential}. Many of these Machine Learning solutions rely on carefully-crafted \textit{features} that encode various aspects of the input SAT instance, e.g. hardness. However, SAT problem instances have quite rich generic and domain-specific structures that are typically lost in the process by most classical feature extraction techniques. This would raise the question of how these structural signals in the input space can be captured and fed to the underlying Machine Learning models. Thanks to new advances in the the field of Representation Learning and particularly Geometric Deep Learning \cite{bronstein2017geometric, wu2019comprehensive}, there have been recent efforts to use Graph Neural Networks (e.g. \cite{li2015gated, defferrard2016convolutional}) to capture the structure of SAT instances -- in particular, the NeuroSAT framework \cite{selsam2018learning}, the Circuit-SAT framework \cite{amizadeh2018learning}, and Recurrent Relational Networks for Sudoku \cite{NIPS2018_7597}. These frameworks have been quite successful in capturing the inherent structure of the SAT instances and \textit{embedding} it into traditional vector spaces that are suitable for Machine Learning models. Nevertheless, in these pure embedding frameworks, it is not clear how the search procedure for a satisfying solution actually works. In other words, there is no (semi) proof of correctness for these methods despite their empirical success. Alternatively, researchers have used deep neural networks within classical search frameworks for tackling combinatorial optimization problems -- e.g. \cite{khalil2017learning}. In these hybrid, neuro-symbolic frameworks, deep learning is typically used to learn optimal search heuristics for a generic classical search algorithm -- e.g. greedy search. While proving correctness in these models is more straightforward, since the search strategy is not being learned, the performance of the resulted models is bounded by the effectiveness of the imposed strategy.  

In this paper, we propose a neural framework for learning CSP (particularly SAT) solvers which effectively belongs to the second category above. In particular, we take the formulation of solving CSPs as probabilistic inference \cite{montanari2007solving, mezard2009information,braunstein2005survey,grover2018streamlining,gableske2013performance} and propose a neural version of it which is capable of learning efficient inference strategy (i.e. the search strategy in this context) for specific problem domains. Our general framework is a design pattern which consists of three main operations: \textit{Propagation}, \textit{Decimation} and \textit{Prediction}, and hence referred as the PDP framework. In general, these operations can be implemented either as fixed algorithms or as trainable neural networks. Generally speaking, PDP can be seen as probabilistic inference in the latent space, and as a result, it is somewhat straightforward to establish how the search strategy in the neural PDP works, unlike pure embedding methods. On the other hand, due to the distributed nature of its decimation component, PDP is \textit{not} restricted by the greedy strategy of the classical decimation process, meaning that it can potentially learn search strategies which are not greedy. And this would distinguishes it from the neuro-symbolic methods that are defined within the greedy strategy. Furthermore, we propose an unsupervised, fully differentiable training mechanism based on \textit{energy minimization} which can train PDP directly toward solving SAT via end-to-end backpropagation. The unsupervised nature of our proposed training mechanism enables PDP to train on (infinite) stream of unlabeled data. Our experimental results show the superiority of the PDP framework compared to both neural and classical solvers. We further show neural PDP also comes close to the state-of-the-art CDCL solver.     
\section{Related Work} \label{sec:related}
Classical Machine Learning has been incorporated in solving combinatorial optimization problems \cite{bengio2018machine} and SAT in particular: from SAT classification \cite{xu2012predicting},
and solver selection \cite{xu2008satzilla}
to
configuration tuning
\cite{haim2009restart, hutter2011sequential, singh2009avatarsat} and branching prediction \cite{liang2016learning,grozea2014can,flint2012perceptron}.

However, more recently, researchers have used Deep Learning to train full-stack solvers. There are two main categories of Deep Learning methodologies proposed recently. In the first category, neural networks are used to embed the input CSP instances into a latent vector space where a predictive model can be trained. \cite{NIPS2018_7597} used Recurrent Relational Networks to train Sudoku solvers. Their framework relies on provided solutions at the training time as opposed to our framework, which is completely unsupervised. \cite{selsam2018learning} proposed to use Graph Neural Networks \cite{li2015gated} to embed CNF instances for the SAT classification problem. They also proposed to use a post-processing clustering approach to decode SAT solutions. In contract, our method is fully unsupervised and is \textit{directly} trained toward solving CSPs. \cite{amizadeh2018learning} proposed a DAG Neural Network to embed logical circuits for solving the Circuit-SAT problem. Their framework is also unsupervised but it is mostly suitable for circuit inputs with DAG structure. \cite{prates2018learning} proposed a convolutional embedding-based method for solving TSP.  

In the second category, neural networks are used to learn useful search heuristics \textit{within} an algorithmic search framework -- typically the greedy search \cite{vinyals2015pointer, bello2016neural, khalil2017learning}, branch-and-bound search \cite{he2014learning} or tree search \cite{li2018combinatorial}. While these methods enjoy a strong inductive bias in learning the optimization algorithm as well as some proof of correctness, their effectiveness is essentially bounded by sub-optimality of the imposed search strategy. Our proposed framework effectively belongs to this category but at the same time, its performance is not bounded by any search strategy.

Our framework can also be seen as learning optimal message passing strategy on probabilistic graphical models. There have been some efforts in this direction \cite{ross2011learning, lin2015deeply, heess2013learning, johnson2016composing,yoon2018inference}, but most are focused on merely message passing, whereas our framework learns both optimal message passing \textit{and} decimation strategies, concurrently.   
\section{Background} \label{sec:bg}

A Constraint Satisfaction Problem, denoted by $\mathbb{CSP}\langle X,C\rangle$, is an instance of combinatorial optimization problem where the goal is to find an \textit{assignment} to a set of $N$ discrete variable $X=\{x_i: i\in 1..N\}$ each defined on a set of discrete values $\boldsymbol{\mathcal{X}}$ such that it satisfies \textit{all} $M$ constraints $C=\{c_a(\boldsymbol{x}_{\partial a}): a \in 1..M\}$. Here, $\partial a$ is a subset of variable indices that constraint $c_a$ depends on; similarly, by $\partial i$, we denote the subset of constraint indices that variable $i$ participates in. Each constraint $c_a:\boldsymbol{\mathcal{X}}^{|\partial a|}\mapsto\{0,1\}$ is a Boolean function that takes value $1$ iff $\boldsymbol{x}_{\partial a}$ satisfies the constraint $c_a$. In this paper, we focus on Boolean Satisfiability problem (SAT) where the variables take values from $\boldsymbol{\mathcal{X}}=\{0,1\}$ and each constraint (or \textit{clause}) is a disjunction of a subset of variables or their negations.

\textbf{CSP as probabilistic inference:} Any CSP instance $\mathbb{CSP}\langle X,C\rangle$ can be represented as a \textit{factor graph} probabilistic graphical model $\mathbb{FG}\langle X,C\rangle$\cite{koller2009probabilistic}. A factor graph $\mathbb{FG}\langle X,C\rangle$ is a bipartite graph where each variable $x_i$ corresponds to a variable node in $\mathbb{FG}$ and each constraint $c_a$ corresponds to a factor node in $\mathbb{FG}$. There is an edge between the $i$-th variable node and the $a$-th factor node if $i\in \partial a$. Then, one may define a measure on FG as:
\begin{equation}
    P(X)=\frac{1}{Z}\prod_{a=1}^M \phi_a(\boldsymbol{x}_{\partial a})\label{eq:measure}
\end{equation}
where $\phi_a$ are the factor functions such that $\phi_a(\boldsymbol{x}_{\partial a}):=\max(c_a(\boldsymbol{x}_{\partial a}), \epsilon)$ for some very small, positive $\epsilon$. $Z$ is the normalization constant. In the special case of SAT, we extend the $\mathbb{FG}$ representation by assigning a binary $e_{ia}\in\{-1, 1\}$ attribute to each edge such that $e_{ia}=-1$ if variable $x_i$ appears negated in the clause $c_a$, and $e_{ia}=1$ otherwise. This way the factor functions take the same functional form (i.e. conjunction) independent of the factor index $a$; that is, $\phi_a(\boldsymbol{x}_{\partial a})=\phi(\boldsymbol{x}_{\partial a}, \boldsymbol{e}_{\partial a})$, where $\boldsymbol{e}_{\partial a}$ are all the edges connected to the $a$-th factor.   

Using this probabilistic formalism, the solutions of the original $\mathbb{CSP}\langle X,C\rangle$ correspond to the modes of $P(X)$. Given $\mathbb{FG}\langle X,C\rangle$, one can compute the marginal distribution of each variable node by doing probabilistic inference on the factor graph using the \textit{Belief Propagation} (BP) algorithm (aka the Sum-Product algorithm) \cite{koller2009probabilistic}. But the actual optimization problem can be solved by computing the \textit{max-marginals} of $P(X)$ via algorithms such as Max-Product, Min-Sum \cite{koller2009probabilistic} and Warning-Propagation \cite{braunstein2005survey}. 

All of the aforementioned algorithms including BP can be seen as special cases of the General Message Passing (GMP) algorithm on factor graphs \cite{mezard2009information}, where the outgoing messages from the graph nodes are computed as a deterministic function of the incoming messages in an iterative fashion. If GMP converges, at the fixed-point, the messages often contain valuable information regarding variable assignments that maximizes the marginal distributions and eventually solve the CSP. In particular, the basic procedure to solve CSPs via probabilistic inference is (1) run a specific GMP algorithm on the factor graph until convergence, (2) based on the incoming fixed-point messages to each variable node, pick the variable with the highest \textit{certainty} regarding a satisfying assignment, (3) set the most certain variable to the corresponding value, simplify the factor graph if possible and repeat the entire process over and over until all variables are set. We refer to this process as \textit{GMP-guided sequential decimation} or decimation for short. The most famous algorithms in this class are BP-guided decimation \cite{montanari2007solving} and  SP-guided decimation, based on the Survey Propagation (SP) algorithm \cite{aurell2005comparing, chavas2005survey}.         

\section{The PDP Framework} \label{sec:pdp}

In order to develop a neural framework for learning to solve CSPs, first we need a suitable \textit{design pattern} that would allow solving CSPs via neural networks. To achieve this, we introduce the \textit{Propagation-Decimation-Prediction} (PDP) framework. The PDP framework can be seen as the generalization of the GMP-guided sequential decimation procedure described in previous section, where certain restrictions are relaxed. In particular:
\begin{itemize}
    \item [(A)] In the GMP-guided sequential decimation, a decimation step is executed only after GMP is converged. We relax this requirement in PDP by interleaving the propagation and the decimation steps.
    \item [(B)] In sequential decimation, at each decimation step, only one variable is fixed (i.e. the one with the highest certainty). But as it is shown in \cite{chavas2005survey}, that does not necessarily need to be the case, and multiple variables can be set concurrently in a fully distributed fashion. We follow this pattern and let the decimation step occur concurrently across the factor graph \textit{without} any centralized selection procedure.
    \item [(C)] In the classical decimation procedure, decimation refers to "fixing" a variable node to a certain value. In PDP, we relax this requirement in two ways: (1) we let the decimation step happen at the message-level (i.e. on the edges) rather than at the variable-level (i.e. on the nodes), and (2) instead of fixing the message on an edge to a certain value, the PDP's decimation step simply intercepts the propagator messages and \textit{transform} them in a \textit{stateful} manner. That is, the decimation step alters the message value in a soft fashion while remembering how it had altered the same message in the previous iterations.   
\end{itemize}
Note that almost all GMP-guided decimation algorithms can be expressed in terms of the PDP design pattern. But on the top of that, the PDP framework offers propagation-decimation mechanisms that cannot be captured by the classical GMP-guided sequential version. For example, in the classical case, the decimation process is greedy by definition; whereas in PDP, that restriction has been lifted.

In what follows, we illustrate the components of the PDP framework in details. Even though our proposed framework is versatile enough to tackle any type of CSP, for the rest of this paper, we focus on the SAT problem.

\subsection{The Messages}
In the PDP framework, at each time step $t$, there are four messages defined on each edge $(i,a)$: the propagator messages in each direction, $\boldsymbol{p}_{i\rightarrow a}^{(t)}$ and $\boldsymbol{p}_{a\rightarrow i}^{(t)}$, and the decimator messages in each direction: $\boldsymbol{d}_{i\rightarrow a}^{(t)}$ and $\boldsymbol{d}_{a\rightarrow i}^{(t)}$. These messages are assumed to be vectors in some latent space $\mathbb{R}^h$.    

\subsection{The Propagation Step}
The propagation step defines how the propagator messages on the edges of the factor graph get updated given the incoming decimator messages from the previous step. In particular for each edge $(i, a)$, we have:
\begin{align}
    \boldsymbol{p}_{i\rightarrow a}^{(t)}&=\boldsymbol{\Psi}_{\theta}\big(\{(\boldsymbol{d}_{b\rightarrow i}^{(t-1)}, e_{bi}): b\in \partial i \setminus a\}\big)\label{eq:propagation1}\\
    \boldsymbol{p}_{a\rightarrow i}^{(t)}&=\boldsymbol{\Psi}_{\gamma}\big(\{(\boldsymbol{d}_{j\rightarrow a}^{(t-1)},e_{aj}): j\in \partial a \setminus i\}\big)\label{eq:propagation2}
\end{align}
where $\boldsymbol{\Psi}_{\theta}$ and $\boldsymbol{\Psi}_{\gamma}$ are general (neural network) \textit{set functions} parametrized by the parameter vectors $\boldsymbol{\theta}$ and $\boldsymbol{\gamma}$, respectively. Similar to \cite{amizadeh2018learning}, in our implementation, we have used Deep Set functions \cite{zaheer2017deep} to model $\boldsymbol{\Psi}_{\theta}$ and $\boldsymbol{\Psi}_{\gamma}$. It should be emphasized that both $\boldsymbol{\Psi}_{\theta}$ and $\boldsymbol{\Psi}_{\gamma}$ are \textit{stateless} functions that compute the outgoing propagator messages merely based on incoming decimator messages and the corresponding edge attributes.

\subsection{The Decimation Step}
As mentioned before, in PDP, the decimation step simply consists of \textit{transforming} the messages generated by the propagation step on each individual edge. Moreover, in principle, the effect of decimation usually goes beyond one iteration, which means that the decimator needs to keep a \textit{memory} of how it transformed the same message in the previous iterations. In other words, the decimation step is inherently \textit{stateful}. Therefore, we define the decimation step as a stateful function that calculates the decimator message on each edge based on the propagator message on the same edge as well as the decimator message at the previous iteration:
\begin{align}
    \boldsymbol{d}_{i\rightarrow a}^{(t)}&=\boldsymbol{\Phi}_{\nu}(\boldsymbol{p}_{i\rightarrow a}^{(t)}, e_{ia}, \boldsymbol{d}_{i\rightarrow a}^{(t-1)})\label{eq:decimation1}\\
    \boldsymbol{d}_{a\rightarrow i}^{(t)}&=\boldsymbol{\Phi}_{\omega}(\boldsymbol{p}_{a\rightarrow i}^{(t)}, e_{ia}, \boldsymbol{d}_{a\rightarrow i}^{(t-1)})\label{eq:decimation2}
\end{align}
Here $\boldsymbol{\Phi}_{\nu}$ and $\boldsymbol{\Phi}_{\omega}$ are recurrent neural network units (e.g. LSTM or GRU \cite{chung2014empirical}) parametrized by parameter vectors $\boldsymbol{\nu}$ and $\boldsymbol{\omega}$, respectively.

\subsection{The Prediction Step}
At any point during the propagation-decimation process, the model can be queried to produce the most likely (soft) assignments for the variable nodes in the factor graph. This is done via the prediction step which produces variable assignments based on the incoming decimator messages to each variable node as well as the corresponding edge attributes; that is,
\begin{equation}
    x_i^{(t)}=\boldsymbol{\Gamma}_{\zeta}\big(\{(\boldsymbol{d}_{b\rightarrow i}^{(t)},e_{bi}): b\in \partial i\}\big)\label{eq:prediction}
\end{equation}
where $\boldsymbol{\Gamma}_{\zeta}$ is another deep set function neural network parametrized by the parameter vector $\boldsymbol{\zeta}$. In the SAT problem, we use the Sigmoid function as the last activation layer of $\boldsymbol{\Gamma}_{\zeta}$ so that we can generate soft assignment for Boolean variables in $(0,1)$ which can be further thresholded at $0.5$ to produce the hard binary assignments.

The tuple $\mathcal{M}=\langle \boldsymbol{\Psi}_{\theta}, \boldsymbol{\Psi}_{\gamma}, \boldsymbol{\Phi}_{\nu}, \boldsymbol{\Phi}_{\omega}, \boldsymbol{\Gamma}_{\zeta}\rangle$ fully specifies the PDP model. Algorithm \ref{alg:pdp} illustrates the interplay of the PDP's three steps in the forward path. Note that at the train time, the forward computation returns back one set of assignments per each iteration step -- i.e. a total of $T_{max}$ sets of soft assignments. This is mainly done for the loss computation purposes, as we will see in the next section. At the test time, however, the iteration loop terminates as soon as a satisfying assignment is found.

Finally note that, at first sight, our proposed PDP framework bears a resemblance to the NeuroSAT model in \cite{selsam2018learning} in the sense that both methods produce latent representations on a bipartite graph via some notion of "message passing". Nevertheless, we would like to emphasize that these methods are fundamentally different. In NeuroSAT, the latent representations are defined per each node and represent the node embeddings of the bipartite representation of CNF, whereas in PDP, the latent vectors represent messages per each \textit{directed} edge. Also, the message passing in NeuroSAT refers to the process proposed in \cite{li2015gated} for graph embedding, whereas in our framework, message passing refers to belief propagation on the factor graph probabilistic graphical model.

\subsection{Parallelization and Batch Replication}
Our implementation of the PDP framework in PyTorch is embarrassingly parallel: every inner loop in Algorithm \ref{alg:pdp} runs concurrently on all the edges/nodes of the input factor graph. Furthermore, parallel processing of multiple input factor graphs in a single batch is quite straightforward in PDP: we simply \textit{concatenate} all the instances present in a batch into one large factor graph and treat the result as one CNF formula with multiple independent clauses. This idea is particularly powerful because it allows us to solve many examples simultaneously on GPU.

Furthermore, batch parallelization can be used to expedite the search in single problem scenarios as well. Note that a CSP can have many solutions and which one of them is found by PDP depends on the random initial message values set in Line 2 of Algorithm \ref{alg:pdp}. Therefore, at the test time, we can \textit{replicate} each single example in the batch multiple times knowing that each replica will be initialized by different random values; then, the iteration loop terminates as soon as the solver finds a solution for at least one of the replicas. This process, which we refer as \textit{batch replication}, enables the solver to simultaneously search different parts of the configuration space for a single problem.     

\begin{algorithm}[tb]
   \caption{The PDP Forward Computation Algorithm}
   \label{alg:pdp}
\begin{algorithmic}[1]
   \STATE {\bfseries Input:} Factor graph $\mathbb{FG}\langle X,C\rangle$, $T_{max}$
    \STATE Randomly initialize $\boldsymbol{p}_{i\rightarrow a}^{(0)}, \boldsymbol{p}_{a\rightarrow i}^{(0)}, \boldsymbol{d}_{i\rightarrow a}^{(0)}, \boldsymbol{d}_{a\rightarrow i}^{(0)}$
   \FOR{$t=1$ {\bfseries to} $T_{max}$}
    \STATE /* Propagation step */
   \FOR{$(i,a)\in \mathbb{FG}\langle X,C\rangle$}
   \STATE Compute $\boldsymbol{p}_{i\rightarrow a}^{(t)}, \boldsymbol{p}_{a\rightarrow i}^{(t)}$ using Eqs. \eqref{eq:propagation1}, \eqref{eq:propagation2}
   \ENDFOR
    \STATE /* Decimation step */
   \FOR{$(i,a)\in \mathbb{FG}\langle X,C\rangle$}
   \STATE Compute $\boldsymbol{d}_{i\rightarrow a}^{(t)}, \boldsymbol{d}_{a\rightarrow i}^{(t)}$ using Eqs. \eqref{eq:decimation1}, \eqref{eq:decimation2}
   \ENDFOR
   \STATE /* Prediction step */
   \FOR{$i=1$ {\bfseries to} $N$}
   \STATE Compute $x_i$ using Eq. \eqref{eq:prediction}
   \ENDFOR
   \STATE $X^{(t)}\gets \{x_i:i \in 1..N\}$
   \IF{Testing = True {\bfseries and} $X^{(t)}$ is SAT}
   \STATE \Return $X^{(t)}$
   \ENDIF
   \ENDFOR
   \STATE \Return $\{X^{(t)}:t \in 1..T_{max}\}$
\end{algorithmic}
\end{algorithm}

\section{Training a PDP SAT Solver} \label{sec:train}

In order to train a PDP model $\mathcal{M}$ to solve SAT problems, we would ideally like to reward the model outputs that have high probability regarding the measure in Eq. \eqref{eq:measure}. As a result, one can train $\mathcal{M}$ by maximizing the probability of the model output w.r.t. the model parameters. Instead, a more numerically stable method aims at minimizing the negative log-probability function known as the \textit{energy function}:
\begin{equation}
    \mathcal{E}(X)=-\log P(X)=\log Z - \sum_{a=1}^M \log\phi(\boldsymbol{x}_{\partial a}, \boldsymbol{e}_{\partial a})\label{eq:energy}
\end{equation}
Nevertheless, $\phi(\cdot)$ is not a differentiable function as we defined it in Section \ref{sec:bg}. Moreover, $\mathcal{M}$ produces soft assignments which cannot be directly fed into $\phi(\cdot)$. To resolve these issues, for training purposes only, we define a differentiable proxy to the original $\phi(\cdot)$. In particular, in the SAT problem, $\phi(\cdot)$ should encode the logical disjunction on soft assignments. One possible differentiable formulation is the \textit{smooth max} function as proposed by \cite{amizadeh2018learning}:
\begin{equation}
    S_{max}(\boldsymbol{x}_{\partial a}, \boldsymbol{e}_{\partial a})=\frac{\sum_{i\in \partial a} e^{\ell(x_i, e_{ia}) / \tau} \ell(x_i, e_{ia}) }{\sum_{i\in \partial a} e^{\ell(x_i, e_{ia}) / \tau}}\label{eq:smax}
\end{equation}
where,
\begin{equation}
    \ell(x_i, e_{ia})=
    \begin{cases}
    x_i & \text{if } e_{ia} = 1\\
    1-x_i & \text{otherwise}
    \end{cases}
\end{equation}
is the \textit{literal function}, and $\tau$ is the temperature parameter. Similar to \cite{amizadeh2018learning}, we start the training at high a temperature and gradually anneal it toward $0$. By doing so, we effectively let $S_{max}(\cdot)$ start off as the arithmetic mean function and gradually turn into the $\max(\cdot)$ function, which is the equivalent of logical disjunction on soft assignments. This is beneficial in the sense that in the beginning of training, we let the gradient signal propagate back via \textit{all} $S_{max}(\cdot)$ inputs equally, which will in turn promote \textit{exploration} in the beginning of training. Note that the output of $S_{max}(\cdot)$ is still a soft assignment even when the temperature is very close to $0$. In order to mimic the behavior of disjunction even further, one can enhance the \textit{contrast} of the $S_{max}(\cdot)$ output. By contrast enhancement, we mean pushing the soft assignment values to the extremes depending on whether they are below $0.5$ or above it. One obvious choice for such transformation is the smooth Step function with transition on $0.5$:
\begin{equation}
    G_{\kappa}(x)=\frac{x^{\kappa}}{x^{\kappa} + (1-x)^{\kappa}}\label{eq:sharp}
\end{equation}
where $\kappa > 1$ is a constant. Using Eqs. \eqref{eq:smax}, \eqref{eq:sharp}, finally we can define our smooth proxy for $\phi(\cdot)$ as:
\begin{equation}
    \phi(\boldsymbol{x}_{\partial a}, \boldsymbol{e}_{\partial a}) = G_{\kappa}\big(S_{max}(\boldsymbol{x}_{\partial a}, \boldsymbol{e}_{\partial a})\big)
\end{equation}
Given this proxy, we define the loss function for the example $\mathbb{FG}\langle X,C\rangle$ as the discounted accumulated energy over $T_{max}$ iterations:
\begin{equation}
    \mathcal{L}_{\lambda}\big(\mathbb{FG}\langle X,C\rangle\big)=\sum_{t=1}^{T_{max}} \lambda^{(T_{max} - t)}\mathcal{E}(X^{(t)})
\end{equation}
where $\{X^{(t)}:t \in 1..T_{max}\}$ is the output of model $\mathcal{M}$ according to Algorithm \ref{alg:pdp}, and $0 < \lambda \leq 1$ is the \textit{discounting factor}. The idea here is the model will be penalized more if it produces non-SAT assignments further down the iteration loop. Finally, we note that using this loss function, our training methodology is completely unsupervised; that is, our framework does \textit{not} need SAT solutions at the training time nor does it need SAT/UNSAT binary labels. And yet, the proposed mechanism \textit{directly} trains the PDP model toward solving SAT. In that respect, our framework is very different from  
the NeuroSAT framework \cite{selsam2018learning} where solver is built indirectly via training a binary SAT classifier.
\section{Experimental Results} \label{sec:exp}

We have compared our PDP framework against three different categories of baselines: (a) the classical probabilistic inference based techniques for SAT solving, (b) the state-of-the-art neural embedding SAT solver and (c) one of the state-of-the-art industrial SAT solvers. In particular, we have experimented with the following methods:

\textbf{SP}: Survey-propagation guided decimation \cite{mezard2009information} is one of the most well-known inference-based algorithms to approach random $k$-SAT problems in the hard SAT region. 

\textbf{Reinforce}: Unlike PDP, the decimation process in SP is sequential. Not only can this slow down the search but also it restricts the search to the greedy strategy. To address these issues, \cite{chavas2005survey} have proposed a fully distributed version of SP called the \textit{Reinforce Algorithm}.

\textbf{NeuroSAT}: NeuroSAT \cite{selsam2018learning} is the state-of-the-art neural-embedding framework based on learning a SAT/UNSAT classifier.

\textbf{Glucose}: Glucose \cite{audemard2018glucose} is the  state-of-the-art Conflict-Driven Clause Learning (CDCL) SAT solver \cite{biere2009conflict} whose basic techniques have become common practice for many modern SAT solvers.

Of these baselines, we have implemented SP and Reinforce via the PDP framework; therefore our neural PDP does not have any unfair advantage in terms of parallelization compared to these methods. NeuroSAT is also highly parallel. All the PDP-based methods as well as NeuroSAT take a maximum iteration number (i.e. $T_{max}$) which controls the timeout budget; in our experiments, we have set $T_{max}=1000$. Glucose, on the other hand, does not have such input parameter; therefore, we have incorporated explicit timeout policy for Glucose for comparison purposes.

\subsection{Software}
Our PyTorch implementation of the PDP framework is open-sourced.\footnote{https://github.com/Microsoft/PDP-Solver} The code base includes different types of PDP-based neural SAT solvers as well as the SP and REINFORCE algorithms, collectively known as SATYR. 

\subsection{General Setup}
Our neural PDP framework for the experiments in this section is configured as follows. The dimension of the message space ($h$) is set to $150$. The deep set functions $\boldsymbol{\Psi}_{\theta}$, $\boldsymbol{\Psi}_{\gamma}$, and 
$\boldsymbol{\Gamma}_{\zeta}$ are implemented according to the formulation in Theorem 2 in \cite{zaheer2017deep}, where $\rho$ and $\phi$ are each 2-layer Perceptrons. The decimator functions $\boldsymbol{\Phi}_{\theta}$ and  $\boldsymbol{\Phi}_{\gamma}$ are implemented as GRU cells. In order to generate soft assignments, the activation function of the last layer of $\boldsymbol{\Gamma}_{\zeta}$ is the Sigmoid function. All the other activation functions are set to the LogSigmoid function. We have used the Adam optimizer with learning rate of $10^{-4}$ and gradient clipping with norm $0.65$ to train our model. We have also enforced weight decay of $10^{-10}$ as well as dropout with rate $0.2$ for regularization. For the NeuroSAT model, we have used the default settings proposed in \cite{selsam2018learning}.

\subsection{Uniform Random $k$-SAT}
Uniform random $k$-SAT problems (where each clause in the input CNF has exactly $k$ literals and is selected \textit{uniformly} from a set of variables) have been studied in depth in Combinatorial Optimization, Statistical Physics and Coding Theory \cite{mezard2009information}. In particular, researchers have rigorously identified four different phases of complexity most random $k$-SAT problems go through as the ratio of clauses to variables (known as $\alpha\equiv M/N$) grows: the replica-symmetric (RS) phase (aka the easy SAT phase), the dynamical phase, the condensation phase and finally the UNSAT phase \cite{krzakala2007gibbs}. While, for example, Belief Propagation (BP) \cite{montanari2007solving} does not really work beyond the RS phase, SP has been proposed to model the two-level uncertainty present in the hard SAT phases, and hence is superior to BP and its variants \cite{aurell2005comparing, mezard2009information}.

\textbf{Training}: Since the training strategy in Section \ref{sec:train} is fully unsupervised, at the training time, we simply generate a stream of unlabeled examples in memory to train a PDP SAT solver. For the training examples, we let the size of each clause in a CNF varies randomly between $2$ and $10$. Also the $\alpha$ value for each CNF varies between $2$ and $10$, while the number of variables $N$ is chosen uniformly between $4$ and $100$. This guarantees that our training stream will contain all ranges of problems in terms of complexity. 
Training of NeuroSAT, on the other hand, is based on training of a supervised SAT/UNSAT classifier which not only does need binary labels, but also imposes a strict training regime to avoid learning superfluous features, as described in \cite{selsam2018learning}. As a result, we have adopted the same regime to generate offline training data for NeuroSAT. Nevertheless, we were not able to train a NeuroSAT classifier for problems with more than $20$ variables, even when we increased the model capacity. therefore, we trained a model with instances of $4$ to $20$ variables instead.     

\begin{figure}[t!]
\includegraphics[width=\columnwidth]{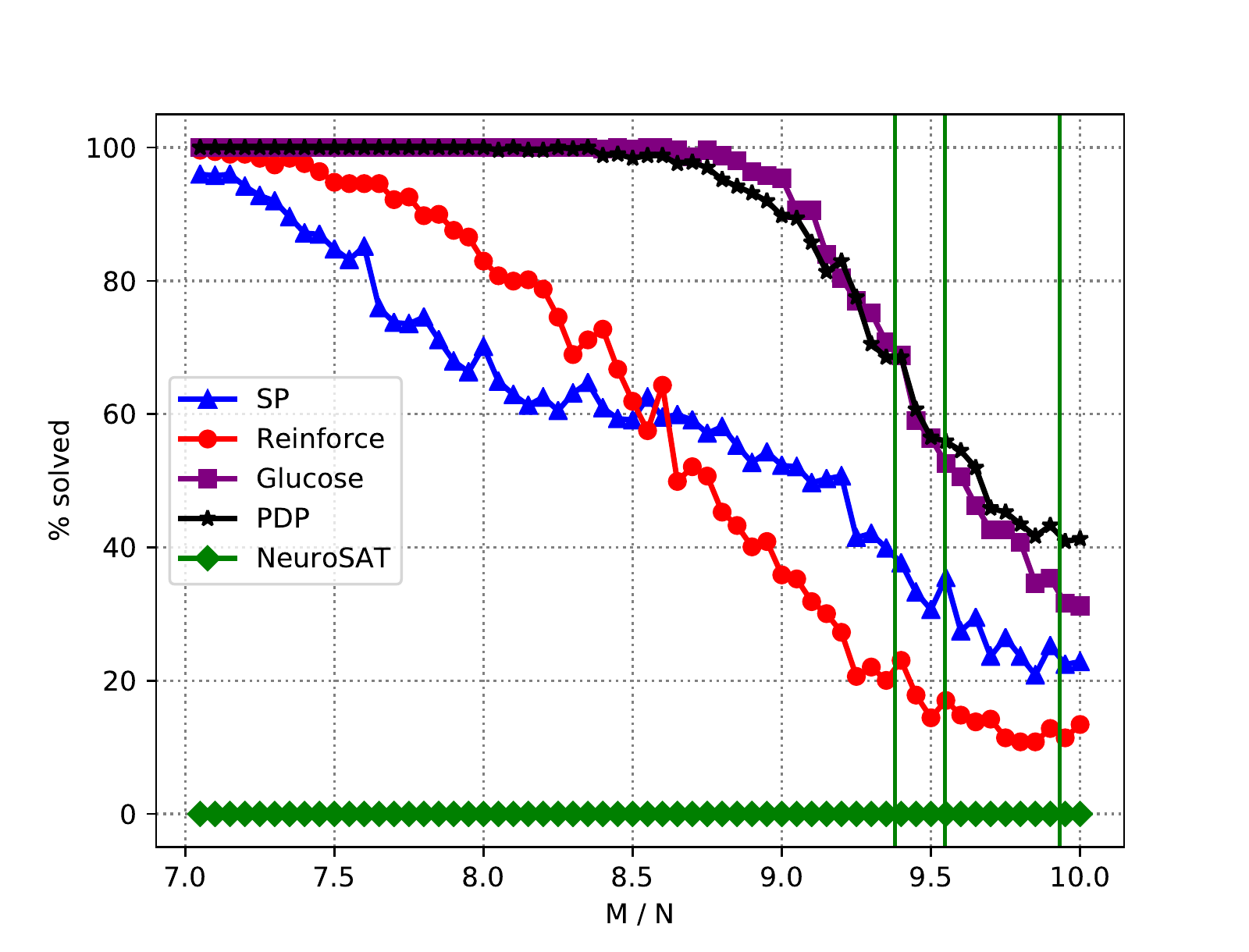}
\centering
\vspace{-0.3in}
\caption{The ratio of the uniform 4-SAT problems solved versus $\alpha$. The green vertical lines indicate the theoretical phase transition thresholds for uniform 4-SAT as $N\rightarrow \infty$.}
\label{fig:4sat}
\vspace{-0.1in}
\end{figure}

\begin{figure}[t!]
\includegraphics[width=\columnwidth]{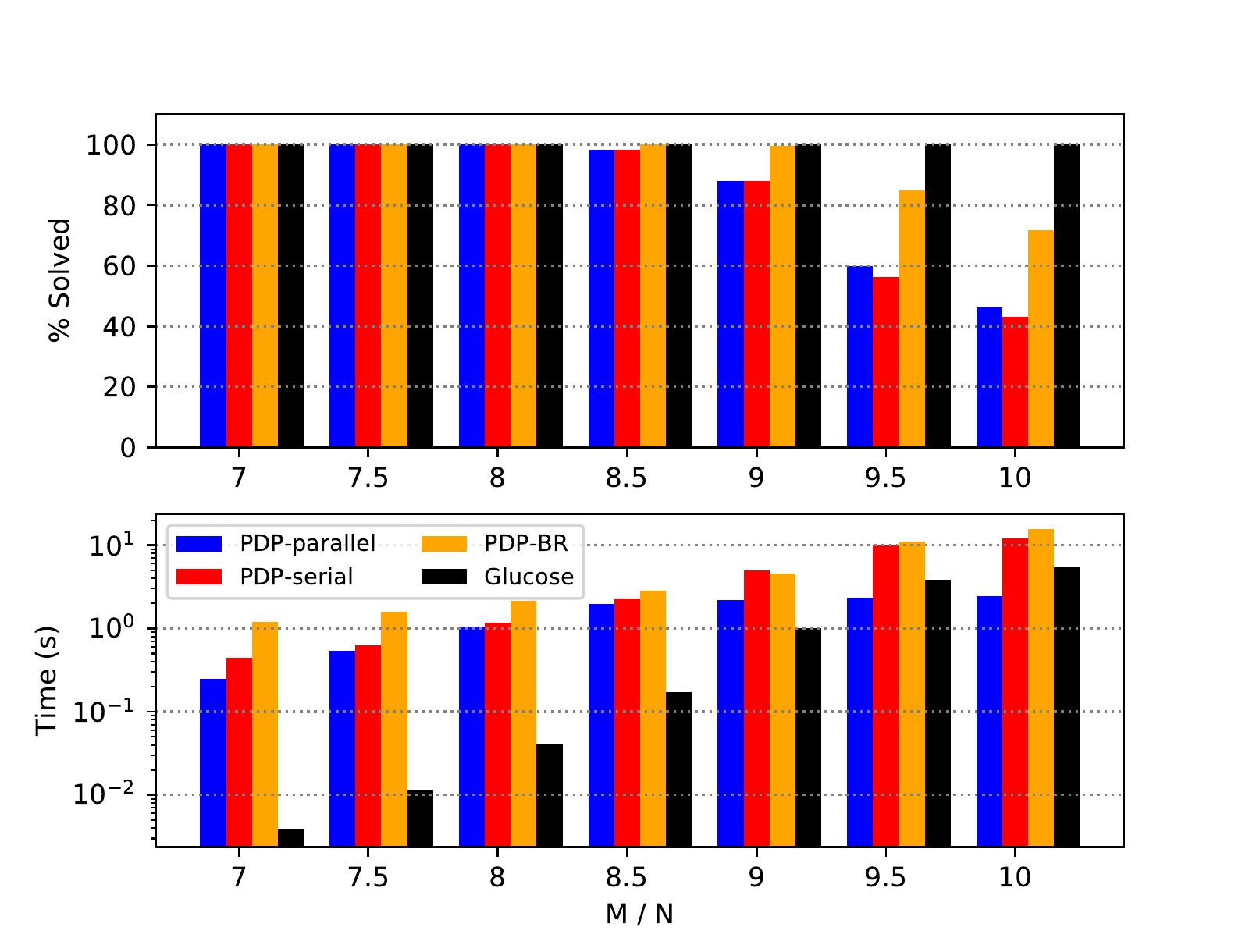}
\centering
\vspace{-0.3in}
\caption{The accuracy and time for Glucose vs. PDP (parallel), PDP (serial) and PDP (serial with batch replication) on uniform 4-SAT. All PDP methods were run up to $1000$ iterations, whereas Glucose was let to solve all the problems without time restriction.}
\label{fig:glucose}
\vspace{-0.1in}
\end{figure}

\textbf{Results}: We have generated test datasets of random satisfiable 4-SAT problems (each with $100$ variables) with various $\alpha$ values. In particular, for any given value of $\alpha$, a dataset of $500$ SAT examples has been generated where all examples in the dataset have roughly the same $\alpha$ value. This process was then repeated for different values of $\alpha$ to cover all the aforementioned phases. Figure \ref{fig:4sat} shows the results of running the models on the 4-SAT test datasets. All the PDP-based models as well as NeuroSAT are run for $T_{max}=1000$ iterations. This would translate to $3$s per example timeout threshold for Glucose. As the plot shows, NeuroSAT model could not solve any of the problems even though its underlying SAT/UNSAT classifier reached a small training error. Unfortunately, we are not able to make any further comment w.r.t. the performance of NeuroSAT since we were not able to train it on larger problems. Our neural PDP method, on the other hand, significantly outperformed the SP and Reinforce algorithms signifying the importance of learning compared to classical (fixed) inference methods. Moreover, within the $3$s timeout budget for Glucose, our framework performs as par with Glucose. This may not seem fair because unlike Glucose, PDP processes multiple examples in a batch each time. To further investigate this, we have let Glucose to go beyond $3$s timeout limit and compared it against the original parallel PDP, its serial version and its serial version augmented by batch replication (each run for $1000$ iterations). Figure \ref{fig:glucose} shows the accuracy and time comparison results. From this plot, we can see that: (1) the serial versions of PDP cannot beat Glucose, (2) Glucose can eventually solve all the problems in the datasets if given opportunity to go beyond $3$s budget, and (3) batch replication significantly improves the accuracy of neural PDP while time overhead is roughly similar to that of serial PDP. The latter observation is crucial because it shows combined with batch replication, PDP can achieve even higher accuracy within limited time budget. This further shows the potential of combining a pure neural framework with classical \textit{restart} mechanism, which is introduced to PDP via batch replication.   

\subsection{Pseudo-Industrial Random $k$-SAT}
As mentioned earlier in this paper, one of the main goals of incorporating Machine Learning in combinatorial optimization is to arrive at generic, data-driven solvers that can adapt to different problem domains. Nevertheless, the neural SAT solvers proposed in the literature so far are mainly \textit{trained} on uniform random $k$-SAT problems. This begs the question whether neural solvers are capable of picking up domain-specific information inherent in a narrow problem distribution. In other words, is one of the main promises of using Machine Learning for combinatorial optimization, namely the adaptability, achievable?    

To address this question, in this section, we go beyond uniform $k$-SAT problems. In particular, we note that despite the heavy focus on uniform $k$-SAT problems in different disciplines, many SAT problems in industrial applications are \textit{not} uniformly distributed but rather exhibit a distinct level of structure. This structure comes in various forms: modularity \cite{ansotegui2012community}, small-world property \cite{walsh1999search}, scale-free property \cite{ansotegui2009structure}, etc. On the other hand, synthetically generating random industrial SAT instances is an active area of research in the SAT community. This is doubly crucial in our scenario where we need abundant amount of training data. In order to generate pseudo-industrial random SAT problems for our experiments, we have generated datasets with high \textit{modularity} structures \cite{newman2006modularity}. This is mainly because it has been shown that real industrial SAT problems are highly modular \cite{ansotegui2012community}. More specifically, we have incorporated the \textit{Community Attachment} (CA) model proposed in \cite{giraldez2016generating} to generate pseudo-industrial modular $k$-SAT datasets.

\textbf{Training}: We have trained two neural PDP models with the same exact architecture and capacity as the one in the previous section. The first model has been trained on a stream of uniformly generated 4-SAT problems with the number of variables ranging from $5$ to $100$. The second model, however, is trained on a stream of random 4-SAT problems generated according to the CA model. Each random example in the CA stream has between $10$ to $20$ communities with modularity factor $Q$ between $0.8$ and $0.9$. We note that the value of $Q$ does not typically go beyond $0.3$ for uniform $k$-SAT problems.  \cite{ansotegui2012community}. 

\begin{figure}[t!]
\includegraphics[width=\columnwidth]{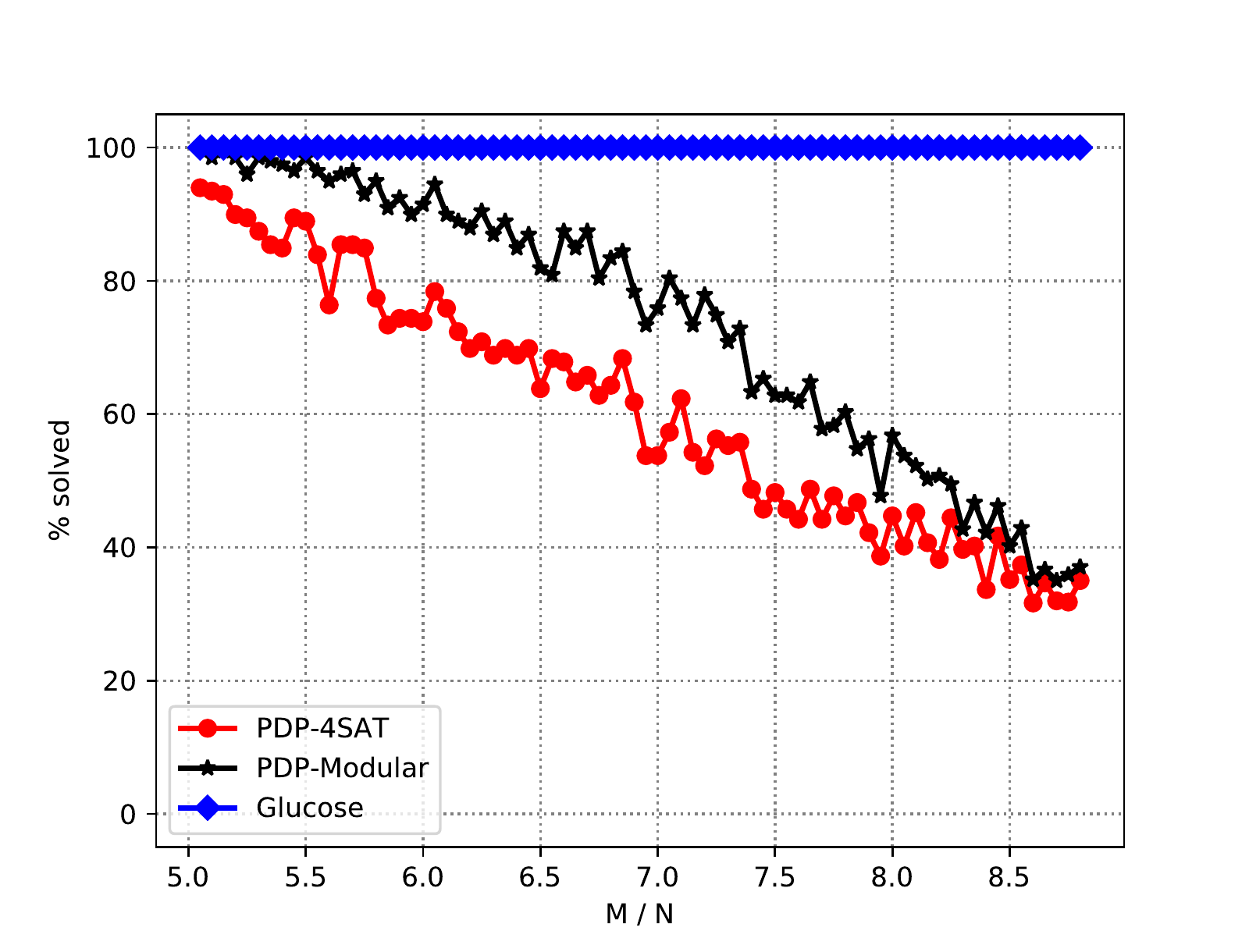}
\centering
\vspace{-0.3in}
\caption{The ratio of the modular 4-SAT problems solved versus $\alpha$ for Glucose and the two PDP solvers trained on uniform and modular random 4-SAT problems, respectively.}
\label{fig:modular}
\vspace{-0.1in}
\end{figure}

\textbf{Results}: We have tested both the trained PDP models as well as Glucose on the datasets of modular 4-SAT problems generated with the same setting as the modular training data for different values of $\alpha$. Figure \ref{fig:modular} shows the performance of these models as $\alpha$ grows. Each point in the plot represents a dataset of $200$ modular examples with the corresponding $\alpha$ value. The PDP models are run for $T_{max}=1000$ iterations which translates to ~$2$s per example timeout budget for Glucose.
As the plot shows, Glucose beats the PDP solvers by solving all the problems for all $\alpha$ values within the time budget. This is somewhat expected as CDCL solvers have been shown to exploit the modular structure \cite{newsham2014impact,giraldez2016generating} and as a result perform very well on real-world industrial problems. However, among the neural PDP models, the one that has been trained on modular 4-SAT problems perform significantly better than the one trained on uniform 4-SAT problems. This is an important result because of the following observation: classical message passing (approximate) inference algorithms (like SP) work reasonably well if the underlying graph is \textit{locally tree-like}. This assumption holds for uniform random $k$-SAT problems as $N\rightarrow \infty$, but does not generally hold for non-uniform random graphs. Nevertheless, the neural PDP can still learn efficient, domain-specific inference strategies.     
This indicates that the neural PDP framework has the potential to adapt to domains beyond uniform random $k$-SAT and be used as a generic yet adaptive solver. 
\section{Conclusions and Future Directions} \label{sec:conclusions}

In this paper, we proposed the neural PDP framework for learning solvers for constraint satisfaction problems. Unlike recent frameworks in the literature which are based on learning efficient embeddings, our framework can be interpreted as a neural extension of probabilistic message passing and inference techniques on graphical models and as such its search strategy can be explained in the probabilistic terms.
Furthermore, we proposed a completely unsupervised training mechanism based on the idea of energy minimization on graphical models to train PDP toward solving SAT. Due to its unsupervised nature, this training mechanism enables us to train PDP on an (infinite) stream of unlabeled problem instances generated in real-time. This is in stark contrast with strict supervised training mechanism taken by some recent frameworks such as NeuroSAT. This also opens the door to the further question of how to generate unlabeled training streams for efficient training in specific domains.   

We also note that PDP as a general design pattern is a powerful idea on its own as it allows many classical message passing algorithms such as SP, BP and Reinforce to be reformulated in terms of a fully parallel framework which can be further run on GPUs. Moreover, the general PDP framework does \textit{not} require its components to be necessarily trainable. This gives rise to \textit{hybrid} models where some components of the model are learned while others are fixed. For example, using PDP, one can easily learn a solver with a SP propagator and a neural decimator. In other words, PDP provides a versatile framework for learning a wide range of neuro-symbolic solvers.      

Since neural PDP actually learns an inference algorithm, it can find multiple solutions for an input CSP depending on its messages' initial values. This further gives rise to the idea of batch replication which in turn introduces the classical notion of restarts in the PDP framework. In this direction, an important future work is to investigate ways of introducing other classical techniques used in CSP solvers, specifically backtracking.      

Our experimental results on $k$-SAT problems showed the effectiveness of the PDP framework compared to classical, neural and state-of-the-art baselines. In particular, we saw that not only did PDP outperform classical baselines but it also came close to the state-of-the-art Glucose solver. Moreover, we showed that PDP can go beyond uniform problems and adapt to pseudo-industrial problem domains with prominent modular structures. Finally, we would like to emphasize that neural PDP is still far from claiming victory over state-of-the-art industrial solvers. However, the fact that our PyTorch prototype could come close to the highly optimized Glucose solver reveals the huge potential in pursuing this direction, especially regarding translating classical techniques used in CDCL solvers into neural frameworks such as PDP. 

\bibliography{ms}

\begin{thebibliography}{52}
\providecommand{\natexlab}[1]{#1}
\providecommand{\url}[1]{\texttt{#1}}
\expandafter\ifx\csname urlstyle\endcsname\relax
  \providecommand{\doi}[1]{doi: #1}\else
  \providecommand{\doi}{doi: \begingroup \urlstyle{rm}\Url}\fi

\bibitem[Amizadeh et~al.(2019)Amizadeh, Matusevych, and
  Weimer]{amizadeh2018learning}
Amizadeh, S., Matusevych, S., and Weimer, M.
\newblock Learning to solve circuit-{SAT}: An unsupervised differentiable
  approach.
\newblock In \emph{International Conference on Learning Representations}, 2019.

\bibitem[Ans{\'o}tegui et~al.(2008)Ans{\'o}tegui, Bonet, Levy, and
  Manya]{ansotegui2008measuring}
Ans{\'o}tegui, C., Bonet, M.~L., Levy, J., and Manya, F.
\newblock Measuring the hardness of sat instances.
\newblock In \emph{AAAI}, volume~8, pp.\  222--228, 2008.

\bibitem[Ans{\'o}tegui et~al.(2009)Ans{\'o}tegui, Bonet, and
  Levy]{ansotegui2009structure}
Ans{\'o}tegui, C., Bonet, M.~L., and Levy, J.
\newblock On the structure of industrial sat instances.
\newblock In \emph{International Conference on Principles and Practice of
  Constraint Programming}, pp.\  127--141. Springer, 2009.

\bibitem[Ans{\'o}tegui et~al.(2012)Ans{\'o}tegui, Gir{\'a}ldez-Cru, and
  Levy]{ansotegui2012community}
Ans{\'o}tegui, C., Gir{\'a}ldez-Cru, J., and Levy, J.
\newblock The community structure of sat formulas.
\newblock In \emph{International Conference on Theory and Applications of
  Satisfiability Testing}, pp.\  410--423. Springer, 2012.

\bibitem[Audemard \& Simon(2018)Audemard and Simon]{audemard2018glucose}
Audemard, G. and Simon, L.
\newblock On the glucose sat solver.
\newblock \emph{International Journal on Artificial Intelligence Tools},
  27\penalty0 (01):\penalty0 1840001, 2018.

\bibitem[Aurell et~al.(2005)Aurell, Gordon, and
  Kirkpatrick]{aurell2005comparing}
Aurell, E., Gordon, U., and Kirkpatrick, S.
\newblock Comparing beliefs, surveys, and random walks.
\newblock In \emph{Advances in Neural Information Processing Systems}, pp.\
  49--56, 2005.

\bibitem[Bello et~al.(2016)Bello, Pham, Le, Norouzi, and
  Bengio]{bello2016neural}
Bello, I., Pham, H., Le, Q.~V., Norouzi, M., and Bengio, S.
\newblock Neural combinatorial optimization with reinforcement learning.
\newblock \emph{arXiv preprint arXiv:1611.09940}, 2016.

\bibitem[Bengio et~al.(2018)Bengio, Lodi, and Prouvost]{bengio2018machine}
Bengio, Y., Lodi, A., and Prouvost, A.
\newblock Machine learning for combinatorial optimization: a methodological
  tour d'horizon.
\newblock \emph{arXiv preprint arXiv:1811.06128}, 2018.

\bibitem[Biere et~al.(2009{\natexlab{a}})Biere, Heule, and van
  Maaren]{biere2009handbook}
Biere, A., Heule, M., and van Maaren, H.
\newblock \emph{Handbook of satisfiability}, volume 185.
\newblock IOS press, 2009{\natexlab{a}}.

\bibitem[Biere et~al.(2009{\natexlab{b}})Biere, Heule, van Maaren, and
  Walsh]{biere2009conflict}
Biere, A., Heule, M., van Maaren, H., and Walsh, T.
\newblock Conflict-driven clause learning sat solvers.
\newblock \emph{Handbook of Satisfiability, Frontiers in Artificial
  Intelligence and Applications}, pp.\  131--153, 2009{\natexlab{b}}.

\bibitem[Braunstein et~al.(2005)Braunstein, M{\'e}zard, and
  Zecchina]{braunstein2005survey}
Braunstein, A., M{\'e}zard, M., and Zecchina, R.
\newblock Survey propagation: An algorithm for satisfiability.
\newblock \emph{Random Structures \& Algorithms}, 27\penalty0 (2):\penalty0
  201--226, 2005.

\bibitem[Bronstein et~al.(2017)Bronstein, Bruna, LeCun, Szlam, and
  Vandergheynst]{bronstein2017geometric}
Bronstein, M.~M., Bruna, J., LeCun, Y., Szlam, A., and Vandergheynst, P.
\newblock Geometric deep learning: going beyond euclidean data.
\newblock \emph{IEEE Signal Processing Magazine}, 34\penalty0 (4):\penalty0
  18--42, 2017.

\bibitem[Chavas et~al.(2005)Chavas, Furtlehner, M{\'e}zard, and
  Zecchina]{chavas2005survey}
Chavas, J., Furtlehner, C., M{\'e}zard, M., and Zecchina, R.
\newblock Survey-propagation decimation through distributed local computations.
\newblock \emph{Journal of Statistical Mechanics: Theory and Experiment},
  2005\penalty0 (11):\penalty0 P11016, 2005.

\bibitem[Chung et~al.(2014)Chung, Gulcehre, Cho, and
  Bengio]{chung2014empirical}
Chung, J., Gulcehre, C., Cho, K., and Bengio, Y.
\newblock Empirical evaluation of gated recurrent neural networks on sequence
  modeling.
\newblock \emph{arXiv preprint arXiv:1412.3555}, 2014.

\bibitem[Defferrard et~al.(2016)Defferrard, Bresson, and
  Vandergheynst]{defferrard2016convolutional}
Defferrard, M., Bresson, X., and Vandergheynst, P.
\newblock Convolutional neural networks on graphs with fast localized spectral
  filtering.
\newblock In \emph{Advances in Neural Information Processing Systems}, pp.\
  3844--3852, 2016.

\bibitem[Flint \& Blaschko(2012)Flint and Blaschko]{flint2012perceptron}
Flint, A. and Blaschko, M.
\newblock Perceptron learning of sat.
\newblock In \emph{Advances in Neural Information Processing Systems}, pp.\
  2771--2779, 2012.

\bibitem[Gableske et~al.(2013)Gableske, M{\"u}elich, and
  Diepold]{gableske2013performance}
Gableske, O., M{\"u}elich, S., and Diepold, D.
\newblock On the performance of cdcl-based message passing inspired decimation
  using $\rho$$\sigma$pmpi.
\newblock In \emph{Pragmatics of SAT Workshop}, 2013.

\bibitem[Gir{\'a}ldez-Cru \& Levy(2016)Gir{\'a}ldez-Cru and
  Levy]{giraldez2016generating}
Gir{\'a}ldez-Cru, J. and Levy, J.
\newblock Generating sat instances with community structure.
\newblock \emph{Artificial Intelligence}, 238:\penalty0 119--134, 2016.

\bibitem[Grover et~al.(2018)Grover, Achim, and Ermon]{grover2018streamlining}
Grover, A., Achim, T., and Ermon, S.
\newblock Streamlining variational inference for constraint satisfaction
  problems.
\newblock In \emph{Advances in Neural Information Processing Systems}, pp.\
  10579--10589, 2018.

\bibitem[Grozea \& Popescu(2014)Grozea and Popescu]{grozea2014can}
Grozea, C. and Popescu, M.
\newblock Can machine learning learn a decision oracle for np problems? a test
  on sat.
\newblock \emph{Fundamenta Informaticae}, 131\penalty0 (3-4):\penalty0
  441--450, 2014.

\bibitem[Haim \& Walsh(2009)Haim and Walsh]{haim2009restart}
Haim, S. and Walsh, T.
\newblock Restart strategy selection using machine learning techniques.
\newblock In \emph{International Conference on Theory and Applications of
  Satisfiability Testing}, pp.\  312--325. Springer, 2009.

\bibitem[He et~al.(2014)He, Daume~III, and Eisner]{he2014learning}
He, H., Daume~III, H., and Eisner, J.~M.
\newblock Learning to search in branch and bound algorithms.
\newblock In \emph{Advances in neural information processing systems}, pp.\
  3293--3301, 2014.

\bibitem[Heess et~al.(2013)Heess, Tarlow, and Winn]{heess2013learning}
Heess, N., Tarlow, D., and Winn, J.
\newblock Learning to pass expectation propagation messages.
\newblock In \emph{Advances in Neural Information Processing Systems}, pp.\
  3219--3227, 2013.

\bibitem[Hutter et~al.(2011)Hutter, Hoos, and
  Leyton-Brown]{hutter2011sequential}
Hutter, F., Hoos, H.~H., and Leyton-Brown, K.
\newblock Sequential model-based optimization for general algorithm
  configuration.
\newblock In \emph{International Conference on Learning and Intelligent
  Optimization}, pp.\  507--523. Springer, 2011.

\bibitem[Johnson et~al.(2016)Johnson, Duvenaud, Wiltschko, Adams, and
  Datta]{johnson2016composing}
Johnson, M., Duvenaud, D.~K., Wiltschko, A., Adams, R.~P., and Datta, S.~R.
\newblock Composing graphical models with neural networks for structured
  representations and fast inference.
\newblock In \emph{Advances in neural information processing systems}, pp.\
  2946--2954, 2016.

\bibitem[Jordi(2015)]{jordi2015classification}
Jordi, L.
\newblock On the classification of industrial sat families.
\newblock In \emph{International Conference of the Catalan Association for
  Artificial Intelligence}, pp.\  163, 2015.

\bibitem[Khalil et~al.(2017)Khalil, Dai, Zhang, Dilkina, and
  Song]{khalil2017learning}
Khalil, E., Dai, H., Zhang, Y., Dilkina, B., and Song, L.
\newblock Learning combinatorial optimization algorithms over graphs.
\newblock In \emph{Advances in Neural Information Processing Systems}, pp.\
  6348--6358, 2017.

\bibitem[Knuth(2015)]{knuth1997art}
Knuth, D.~E.
\newblock \emph{The art of computer programming, Volume 4, Fascicle 6:
  Satisfiability}, volume~4.
\newblock Addison-Wesley Professional, 2015.

\bibitem[Koller et~al.(2009)Koller, Friedman, and
  Bach]{koller2009probabilistic}
Koller, D., Friedman, N., and Bach, F.
\newblock \emph{Probabilistic graphical models: principles and techniques}.
\newblock MIT press, 2009.

\bibitem[Krzaka{\l}a et~al.(2007)Krzaka{\l}a, Montanari, Ricci-Tersenghi,
  Semerjian, and Zdeborov{\'a}]{krzakala2007gibbs}
Krzaka{\l}a, F., Montanari, A., Ricci-Tersenghi, F., Semerjian, G., and
  Zdeborov{\'a}, L.
\newblock Gibbs states and the set of solutions of random constraint
  satisfaction problems.
\newblock \emph{Proceedings of the National Academy of Sciences}, 104\penalty0
  (25):\penalty0 10318--10323, 2007.

\bibitem[Kumar(1992)]{kumar1992algorithms}
Kumar, V.
\newblock Algorithms for constraint-satisfaction problems: A survey.
\newblock \emph{AI magazine}, 13\penalty0 (1):\penalty0 32, 1992.

\bibitem[Li et~al.(2015)Li, Tarlow, Brockschmidt, and Zemel]{li2015gated}
Li, Y., Tarlow, D., Brockschmidt, M., and Zemel, R.
\newblock Gated graph sequence neural networks.
\newblock \emph{arXiv preprint arXiv:1511.05493}, 2015.

\bibitem[Li et~al.(2018)Li, Chen, and Koltun]{li2018combinatorial}
Li, Z., Chen, Q., and Koltun, V.
\newblock Combinatorial optimization with graph convolutional networks and
  guided tree search.
\newblock In \emph{Advances in Neural Information Processing Systems}, pp.\
  537--546, 2018.

\bibitem[Liang et~al.(2016)Liang, Ganesh, Poupart, and
  Czarnecki]{liang2016learning}
Liang, J.~H., Ganesh, V., Poupart, P., and Czarnecki, K.
\newblock Learning rate based branching heuristic for sat solvers.
\newblock In \emph{International Conference on Theory and Applications of
  Satisfiability Testing}, pp.\  123--140. Springer, 2016.

\bibitem[Lin et~al.(2015)Lin, Shen, Reid, and van~den Hengel]{lin2015deeply}
Lin, G., Shen, C., Reid, I., and van~den Hengel, A.
\newblock Deeply learning the messages in message passing inference.
\newblock In \emph{Advances in Neural Information Processing Systems}, pp.\
  361--369, 2015.

\bibitem[Mezard \& Montanari(2009)Mezard and Montanari]{mezard2009information}
Mezard, M. and Montanari, A.
\newblock \emph{Information, physics, and computation}.
\newblock Oxford University Press, 2009.

\bibitem[Montanari et~al.(2007)Montanari, Ricci-Tersenghi, and
  Semerjian]{montanari2007solving}
Montanari, A., Ricci-Tersenghi, F., and Semerjian, G.
\newblock Solving constraint satisfaction problems through belief
  propagation-guided decimation.
\newblock \emph{arXiv preprint arXiv:0709.1667}, 2007.

\bibitem[Newman(2006)]{newman2006modularity}
Newman, M.~E.
\newblock Modularity and community structure in networks.
\newblock \emph{Proceedings of the national academy of sciences}, 103\penalty0
  (23):\penalty0 8577--8582, 2006.

\bibitem[Newsham et~al.(2014)Newsham, Ganesh, Fischmeister, Audemard, and
  Simon]{newsham2014impact}
Newsham, Z., Ganesh, V., Fischmeister, S., Audemard, G., and Simon, L.
\newblock Impact of community structure on sat solver performance.
\newblock In \emph{International Conference on Theory and Applications of
  Satisfiability Testing}, pp.\  252--268. Springer, 2014.

\bibitem[Nudelman et~al.(2004)Nudelman, Leyton-Brown, Hoos, Devkar, and
  Shoham]{nudelman2004understanding}
Nudelman, E., Leyton-Brown, K., Hoos, H.~H., Devkar, A., and Shoham, Y.
\newblock Understanding random sat: Beyond the clauses-to-variables ratio.
\newblock In \emph{International Conference on Principles and Practice of
  Constraint Programming}, pp.\  438--452. Springer, 2004.

\bibitem[Palm et~al.(2018)Palm, Paquet, and Winther]{NIPS2018_7597}
Palm, R., Paquet, U., and Winther, O.
\newblock Recurrent relational networks.
\newblock In Bengio, S., Wallach, H., Larochelle, H., Grauman, K.,
  Cesa-Bianchi, N., and Garnett, R. (eds.), \emph{Advances in Neural
  Information Processing Systems 31}, pp.\  3372--3382. Curran Associates,
  Inc., 2018.

\bibitem[Prates et~al.(2018)Prates, Avelar, Lemos, Lamb, and
  Vardi]{prates2018learning}
Prates, M.~O., Avelar, P.~H., Lemos, H., Lamb, L., and Vardi, M.
\newblock Learning to solve np-complete problems-a graph neural network for the
  decision tsp.
\newblock \emph{arXiv preprint arXiv:1809.02721}, 2018.

\bibitem[Ross et~al.(2011)Ross, Munoz, Hebert, and Bagnell]{ross2011learning}
Ross, S., Munoz, D., Hebert, M., and Bagnell, J.~A.
\newblock Learning message-passing inference machines for structured
  prediction.
\newblock In \emph{Computer Vision and Pattern Recognition (CVPR), 2011 IEEE
  Conference on}, pp.\  2737--2744. IEEE, 2011.

\bibitem[Selsam et~al.(2019)Selsam, Lamm, Bunz, Liang, de~Moura, and
  Dill]{selsam2018learning}
Selsam, D., Lamm, M., Bunz, B., Liang, P., de~Moura, L., and Dill, D.~L.
\newblock Learning a sat solver from single-bit supervision.
\newblock In \emph{International Conference on Learning Representations}, 2019.

\bibitem[Singh et~al.(2009)Singh, Near, Ganesh, and Rinard]{singh2009avatarsat}
Singh, R., Near, J.~P., Ganesh, V., and Rinard, M.
\newblock Avatarsat: An auto-tuning boolean sat solver.
\newblock 2009.

\bibitem[Vinyals et~al.(2015)Vinyals, Fortunato, and
  Jaitly]{vinyals2015pointer}
Vinyals, O., Fortunato, M., and Jaitly, N.
\newblock Pointer networks.
\newblock In \emph{Advances in Neural Information Processing Systems}, pp.\
  2692--2700, 2015.

\bibitem[Walsh et~al.(1999)]{walsh1999search}
Walsh, T. et~al.
\newblock Search in a small world.
\newblock In \emph{Ijcai}, volume~99, pp.\  1172--1177. Citeseer, 1999.

\bibitem[Wu et~al.(2019)Wu, Pan, Chen, Long, Zhang, and
  Yu]{wu2019comprehensive}
Wu, Z., Pan, S., Chen, F., Long, G., Zhang, C., and Yu, P.~S.
\newblock A comprehensive survey on graph neural networks.
\newblock \emph{arXiv preprint arXiv:1901.00596}, 2019.

\bibitem[Xu et~al.(2008)Xu, Hutter, Hoos, and Leyton-Brown]{xu2008satzilla}
Xu, L., Hutter, F., Hoos, H.~H., and Leyton-Brown, K.
\newblock Satzilla: portfolio-based algorithm selection for sat.
\newblock \emph{Journal of artificial intelligence research}, 32:\penalty0
  565--606, 2008.

\bibitem[Xu et~al.(2012)Xu, Hoos, and Leyton-Brown]{xu2012predicting}
Xu, L., Hoos, H.~H., and Leyton-Brown, K.
\newblock Predicting satisfiability at the phase transition.
\newblock In \emph{AAAI}, 2012.

\bibitem[Yoon et~al.(2018)Yoon, Liao, Xiong, Zhang, Fetaya, Urtasun, Zemel, and
  Pitkow]{yoon2018inference}
Yoon, K., Liao, R., Xiong, Y., Zhang, L., Fetaya, E., Urtasun, R., Zemel, R.,
  and Pitkow, X.
\newblock Inference in probabilistic graphical models by graph neural networks.
\newblock \emph{arXiv preprint arXiv:1803.07710}, 2018.

\bibitem[Zaheer et~al.(2017)Zaheer, Kottur, Ravanbakhsh, Poczos, Salakhutdinov,
  and Smola]{zaheer2017deep}
Zaheer, M., Kottur, S., Ravanbakhsh, S., Poczos, B., Salakhutdinov, R.~R., and
  Smola, A.~J.
\newblock Deep sets.
\newblock In \emph{Advances in Neural Information Processing Systems}, pp.\
  3391--3401, 2017.

\end{thebibliography}
\bibliographystyle{icml2019}
\end{document}